\documentclass[runningheads]{llncs}
\usepackage[width=122mm,left=12mm,paperwidth=146mm,height=193mm,top=12mm,paperheight=217mm]{geometry}
\usepackage{times}
\usepackage{epsfig}
\usepackage{graphicx}
\usepackage{xspace}
\usepackage{amsmath}
\usepackage{amssymb}
\usepackage{lipsum}
\usepackage{xspace}
\usepackage{amscd}
\usepackage{algorithm}
\usepackage{algpseudocode}
\usepackage{changepage}
\usepackage{pgfplots}
\usepackage{pgfkeys}
\usepackage{xparse}
\usepackage{multirow}
\usepackage{epstopdf}
\usepackage{array}
\usepackage[font={small}]{caption}
\usepackage{url}
\usepackage{adjustbox}

\usepackage[pagebackref=true,breaklinks=true,letterpaper=true,colorlinks,bookmarks=false]{hyperref}

\newcommand{\real}{\mathbb{R}}
\newcommand{\ba}{\mathbf{a}}
\newcommand{\bb}{\mathbf{b}}

\newcommand{\bff}{\mathbf{f}}

\newcommand{\bp}{\mathbf{p}}

\newcommand{\bx}{\mathbf{x}}

\newcommand{\archsmall}{DetNet-S\xspace}
\newcommand{\archbig}{DetNet-L\xspace}

\newcommand{\detnet}{\textsc{DetNet}}
\newcommand{\rotnet}{\textsc{RotNet}}

\newcolumntype{R}[1]{>{\raggedleft\arraybackslash}m{#1}}
\newcolumntype{C}[1]{>{\centering\arraybackslash}m{#1}}

\pgfplotsset{compat=newest}
\pgfplotsset{
	tick label style={font=\tiny},
	label style={font=\scriptsize},
	legend style={font=\scriptsize},
	title style={font=\scriptsize}}
\newlength\figureheight
\newlength\figurewidth

\newenvironment{customlegend}[1][]{%
	\begingroup
	\csname pgfplots@init@cleared@structures\endcsname
	\pgfplotsset{#1}%
}{%
\csname pgfplots@createlegend\endcsname
\endgroup
}%

\def\addlegendimage{\csname pgfplots@addlegendimage\endcsname}

\NewDocumentCommand{\rot}{O{45} O{1em} m}{\makebox[#2][c]{\rotatebox{#1}{#3}}}%

\newcommand{\mygraphic}[1]{\includegraphics[height=#1]{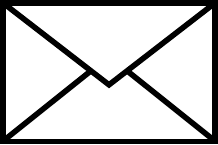}}
\newcommand{\myenv}{\raisebox{0.2pt}{\mygraphic{.45em}}}

\graphicspath{{./figures/}}
\def\ECCV16SubNumber{4}
\title{Learning Covariant Feature Detectors}
\titlerunning{Learning Covariant Feature Detectors}
\authorrunning{Lenc, Vedaldi}
\author{Karel Lenc and Andrea Vedaldi}
\institute{Department of Engineering Science, University of Oxford \\
	\myenv~~\email{karel@robots.ox.ac.uk} \hspace{1em}  \email{vedaldi@robots.ox.ac.uk}}

\begin{document}
\pagestyle{headings}
\mainmatter
\maketitle

\begin{abstract}
Local covariant feature detection, namely the problem of extracting viewpoint invariant features from images, has so far largely resisted the application of machine learning techniques. In this paper, we propose the first fully general  formulation for learning local covariant feature detectors. We propose to cast detection as a regression problem, enabling the use of powerful regressors such as deep neural networks. We then derive a \emph{covariance constraint} that can be used to automatically learn which visual structures provide stable anchors for local feature detection. We support these ideas theoretically, proposing a novel analysis of local features in term of geometric transformations, and we show that all common and many uncommon detectors can be derived in this framework. Finally, we present empirical results on translation and rotation covariant detectors on standard feature benchmarks, showing the power and flexibility of the framework.
\end{abstract}

\section{Introduction}\label{s:intro}

\begin{figure}[t]
\centering
\includegraphics[width=0.75\linewidth]{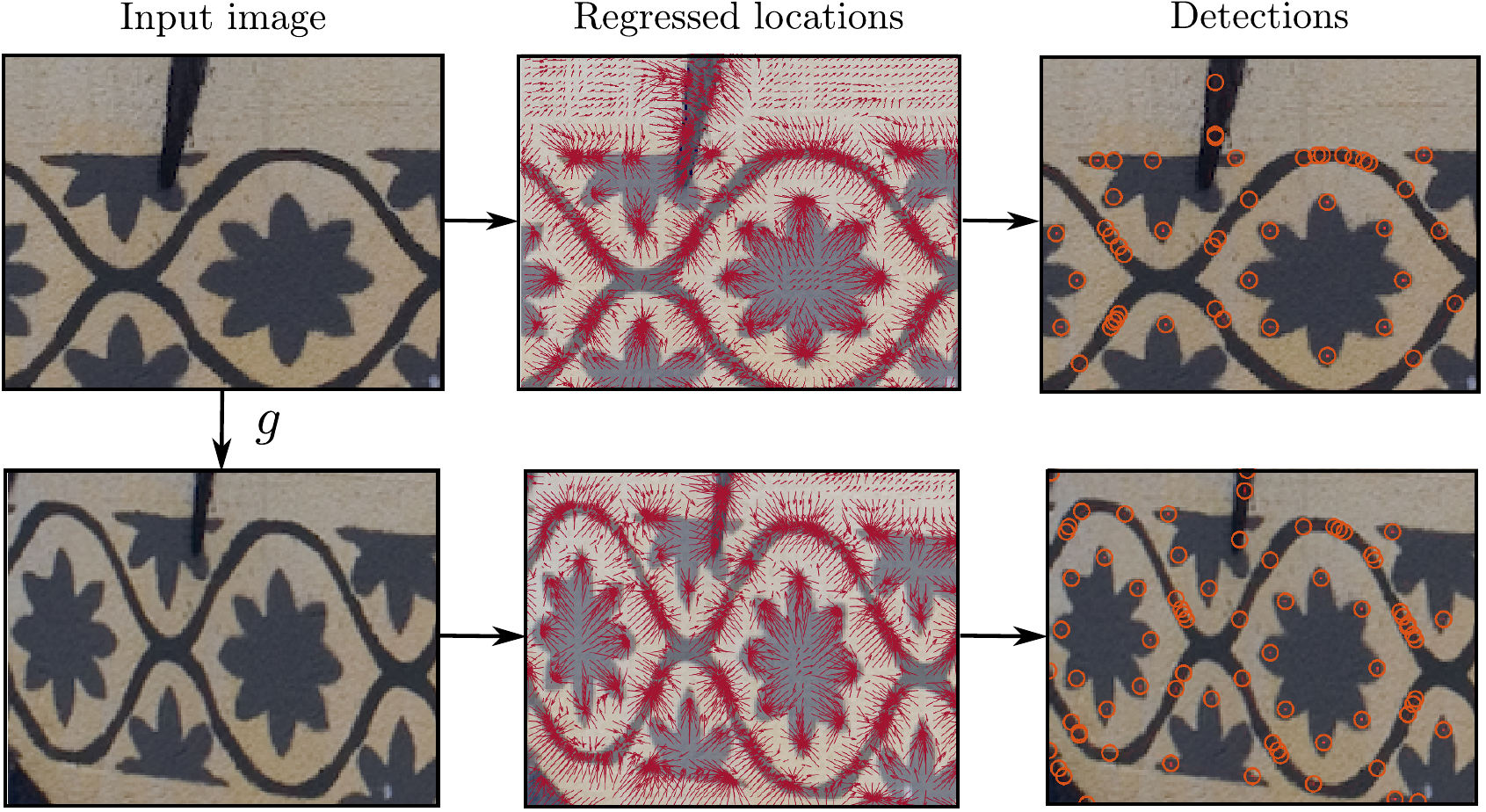}
\caption{\emph{Detection by regression.} We train a neural network $\phi$ that, given a patch $\bx|_p$ around each pixel $p$ in an image, produces a displacement vector $h_p = \phi(\bx|_p)$ pointing to the nearest feature location (middle column). Displacements from nearby pixels are then pooled to detect features (right column). The neural network is trained in order to be covariant with transformations $g$ of the image (bottom row). Best viewed on screen. Image data from \cite{cordes11increasing}. \label{f:spalsh}}
\end{figure}

Image matching, i.e.\ the problem of establishing point correspondences between two images of the same scene, is central to computer vision. In the past two decades, this problem stimulated the creation of numerous \emph{viewpoint invariant local feature detectors}. These were also adopted in problems such as large scale image retrieval and object category recognition, as a general-purpose image representations. More recently, however, deep learning has replaced local features as the preferred method to construct image representations; in fact, the most recent works on local feature descriptors are now based on deep learning~\cite{han15matchnet:,zbontar15computing}. 

Differently from \emph{descriptors}, the problem of constructing local feature \emph{detectors} has so far largely resisted machine learning. The goal of a detector is to extract stable local features from images, which is an essential step in any matching algorithm based on sparse features. It may be surprising that machine learning has not been very successful at this task given that it has proved very useful in many other detection problems. We believe that the reason is the difficulty of devising a learning formulation for viewpoint invariant features.

To clarify this difficulty, note that the fundamental aim of a local feature detector is to extract the same features from images regardless of effects such as viewpoint changes. In computer vision, this behavior is more formally called~\emph{covariant detection}. Handcrafted detectors achieve it by anchoring features to image structures, such as corners or blobs, that are preserved under a viewpoint change. However, there is no \emph{a--priori} list of what visual structures constitute useful anchors. Thus, an algorithm must not only learn the appearance of the anchors, but needs to determine what anchors are in the first place. In other words, the challenge is to learn simultaneously a detector together with the detection targets.

In this paper we propose a method to address this challenge. Our first contribution is to introduce a novel \emph{learning formulation for covariant detectors} (Sect.~\ref{s:method}). This is based on two ideas: i) defining an objective function in term of a \emph{covariance constraint} which is anchor-agnostic (Sect.~\ref{s:covconst}) and ii) formulating detection as a \emph{regression problem}, which allows to use powerful regressors such as deep networks for this task (Fig.~\ref{f:spalsh}).

Our second contribution is to support this approach theoretically. We show how covariant feature detectors are best understood and manipulated in term of image transformations (Sect.~\ref{s:beyond}). Then, we show that, geometrically, different detector types can be characterized by which transformations they are covariant with and, among those, which ones they fix and which they leave undetermined (Sect.~\ref{s:partial}). We then show that this formulation encompasses all common and many uncommon detector types and allows to derive a covariance constraint for each one of them (Sect.~\ref{s:taxonomy}).

Our last contribution is to validate this approach empirically. We do so by first discussing several important implementation details (Sect.~\ref{s:implementation}), and then by training and assessing two different detector types, comparing them to off-the-shelf detectors (Sect.~\ref{s:experiments}). Finally, we discuss future extensions (Sect.~\ref{s:discussion}).

\subsection{Related work}\label{s:related}

\sloppy

Covariant detectors differ by the type of features that they extract: points~\cite{harris88combined,lindeberg94scale-space,schmid97local,dufournaud99matching}, circles~\cite{lindeberg98feature,lowe99object,mikolajczyk01harris-laplace}, or ellipses~\cite{lindeberg94shape-adapted,tuytelaars00wide,baumberg00reliable,schaffalitzky01viewpoint,matas02local,mikolajczyk02affine}. In turn, the type of feature determines which class of transformations that they can handle: Euclidean transformations, similarities, and affinities.

Another differentiating factor is the type of visual structures used as anchors. For instance, early approaches used corners extracted from an analysis of image edglets~\cite{rosenfeld73angle,freeman77a-corner-finding,sankar78a-parallel}. These were soon surpassed by methods that extracted corners and other anchors using operators of the image intensity such as the \emph{Hessian of Gaussian}~\cite{beaudet78rotationally} or the \emph{structure tensor}~\cite{forstner86a-feature,harris88combined,zuliani05mathematical} and its generalizations~\cite{triggs04detecting}. In order to handle transformations more complex than translations and rotations, scale selection methods using the \emph{Laplacian/Difference of Gaussian} operator (L/DoG) were introduced~\cite{lowe99object,mikolajczyk01harris-laplace}, and further extended with \emph{affine adaptation}~\cite{baumberg00reliable,mikolajczyk02affine} to handle full affine transformations. While these are probably the best known detectors, several other approaches were explored as well, including parametric feature models~\cite{guiducci88corner,rohr92recognizing} and using self-dissimilarity~\cite{smith95SUSAN,kadir01saliency}.
 
All detectors discussed so far are \emph{handcrafted}. Learning has been mostly limited to the case in which detection anchors are defined \emph{a-priori}, either by manual labelling~\cite{kienzle06learning} or as the output of a pre-existing handcrafted detetctor~\cite{dias95a-neural,rosten06machine,sochman09learning,holzer12learning}, with the goal of accelerating detection. Closer to our aim, \cite{rosten10faster} use simulated annealing to optimise the parameters of their FAST detector for repeatability. To the best of our knowledge, the only line of work that attempted to learn repeatable anchors from scratch  is the one of~ \cite{trujillo06synthesis,olague11evolutionary-computer-assisted}, who did so using genetic programming; however, their approach is much more limited than ours, focusing only on the repeatability of corner points.

More recently, \cite{Yi16Learning} learns to estimate the orientation of feature points using deep learning. Contrary to our approach, the loss function is defined on top of the local image feature descriptors and is limited to estimating the rotation of keypoints. The work of~\cite{Zagoruyko2015,paulin2015local,simo2015discriminative} also use Siamese deep learning architectures for local features, but for local image feature \emph{description}, whereas we use them for feature \emph{detection}.

\fussy
\section{Method}\label{s:method}

We first introduce our method in a special case, namely in learning a basic corner detector (Sect.~\ref{s:covconst}), and then we extend it to general covariant features (Sect.~\ref{s:beyond} and~\ref{s:partial}). Finally, we show how the theory applies to concrete examples of detectors (Sect.~\ref{s:taxonomy}).

\subsection{The covariance constraint}\label{s:covconst}

Let $\bx$ be an image and let $T\bx$ be its version translated by $T\in\mathbb{R}^2$ pixels. A corner detector extracts from $\bx$ a (small) collection of points $\bff\in\mathbb{R}^2$. The detector is said to be covariant if, when applied to the translated image $T\bx$, it returns the translated points $\bff + T$. Most covariant detectors work by anchoring features to image structures that, such as corners, are preserved under transformation. A challenge in defining anchors is that these must be general enough to be found in most images and at the same time sufficiently distinctive to achieve covariance.

Anchor extraction is usually formulated as a \emph{selection problem} by finding the features that maximize a handcrafted figure of merit such as Harris' cornerness, the Laplacian of Gaussian, or the Hessian of Gaussian. This indirect construction makes learning anchors difficult. As a solution, we propose to regard feature detection not as a selection problem but as a \emph{regression one}. Thus the goal is to learn a function $\psi : \bx \mapsto \bff$ that directly maps an image (patch\footnote{As the function $\psi$ needs to be location invariant it can be applied in a sliding window manner. Therefore $x$ can be a single patch which represents its perception field.}) $\bx$ to a corner $\bff$. The key advantage is that this function can be implemented by any regression method, including a deep neural network.

This leaves the problem of defining a learning objective. This would be easy if we had example anchors annotated in the data; however, our aim is to discover useful anchors automatically. Thus, we propose to \emph{use covariance itself as a learning objective}. This is formally captured by the \emph{covariance constraint} $\psi(T\bx) = T + \psi(\bx)$. A corresponding learning objective can be formulated as follows:
\begin{equation}
	\label{e:simple-loss}
	 \min_{\psi} \frac{1}{n} \sum_{i=1}^n \| \psi(T_i\bx_i) - \psi(\bx_i) - T_i\|^2
\end{equation}
where $(\bx_i,T_i)$ are example patches and transformations and the optimization is over the parameters of the regressor $\psi$ (e.g. the filter weights in a deep neural network).

\subsection{Beyond corners}\label{s:beyond}

\begin{figure}[t]
\begin{center}
	\includegraphics[height=0.13\textheight]{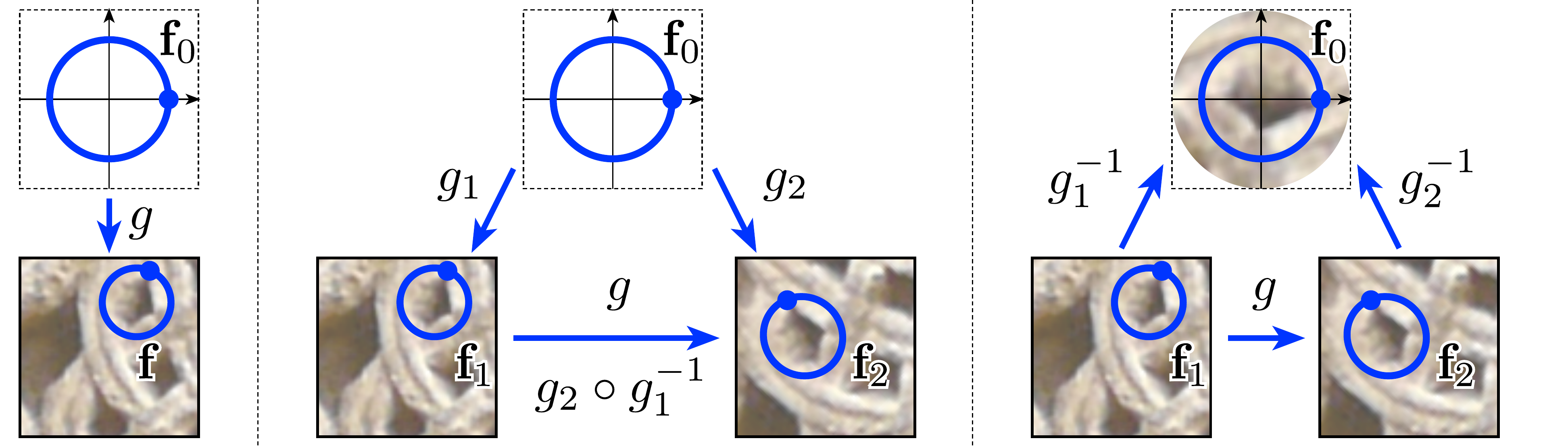}
\end{center}
\caption{\textbf{Left}: an oriented circular frame $\bff = g\bff_0$ is obtained as a unique similarity transformation $g\in G$ of the canonical frame $\bff_0$, where the orientation is represented by the dot. Concretely, this could be the output of the SIFT detector after orientation assignment. \textbf{Middle}: the detector finds feature frames $\bff_i = g_i\bff_0, g_i = \phi(\bx_i)$ in images $\bx_1$ and $\bx_2$ respectively due to covariance, matching the features allows to recover the underlying image transformation $\bx_2 = g \bx_1$ as $g=g_2 \circ g_1^{-1}$. \textbf{Right}: equivalently, then inverse transformations $g_i^{-1}$ normalize the images, resulting in the same canonical view.}\label{f:detector}
\end{figure}

This section provides a first generalization of the construction above. While simple detectors such as Harris extract 2D points $\bff$ in correspondence of corners, others such as SIFT extract circles in correspondence of blobs, and others again extract even more complex features such as oriented circles (e.g.\ SIFT with orientation assignment), ellipses (e.g.\ Harris-Affine), oriented ellipses (e.g.\ Harris-Affine with orientation assignment), etc. In general, due to their role in fixing image transformations, we will call the extracted shapes $\bff\in\mathcal{F}$ \emph{feature frames}.

The detector is thus a function $\psi : \mathcal{X} \rightarrow \mathcal{F},\ \bx \mapsto \bff$ mapping an image patch $\bx$ to a corresponding feature frame $\bff$. We say that the detector is \emph{covariant with a group of transformations}\footnote{Here, a group of transformation $(G,\circ)$ is a set of functions $g,h:\mathbb{R}^2 \rightarrow\mathbb{R}^2$ together with composition $g \circ h\in G$ as group operation. Composition is associative; furthermore, $G$ contains the identity transformation $1$ and the inverse $g^{-1}$ of each of its elements $g\in G$.} $g\in G$ (e.g.\ similarity or affine)  when
\begin{equation}\label{e:covariance}
 \forall \bx\in\mathcal{X}, g\in G: \quad  \psi(g\bx) = g\psi(\bx)
\end{equation}
where $g\bff$ is the transformed frame and $g\bx$ is the warped image.\footnote{The action $g\bx$ of the transformation $g$ on the image $\bx$ is to warp it: $(g\bx)(u,v) = \bx(g^{-1}(u,v))$}

Working with feature frames is intuitive, but cumbersome and not very flexible. A much better approach is to drop frames altogether and replace them with corresponding transformations. For instance, in SIFT with orientation assignment all possible oriented circles $\bff$ can be expressed uniquely as a similarity $g\bff_0$ of a fixed oriented circle $\bff_0$ (Fig.~\ref{f:detector}.left). Hence, instead of talking about oriented circles $\bff$, we can equivalently talk about similarities $g$. Likewise, in the case of the Harris' corner detector, all possible 2D points $\bff$ can be expressed as translations $T + \bff_0$ of the origin $\bff_0$, and so we can talk about translations $T$ instead of points $\bff$.

To generalize this idea, we say that that a class of frames $\mathcal{F}$ \emph{resolves} a group of transformations $G$ when, given a fixed \emph{canonical frame} $\bff_0\in\mathcal{F}$, all frames are \emph{uniquely generated} from it by the action of $G$:
\[
  \mathcal{F} = G \bff_0 = \{ g\bff_0 : g \in G \}
  \quad\text{and}\quad
  \forall g,h\in G: \  g\bff_0 = h \bff_0 \ \Rightarrow\  g = h \text{\ (uniqueness)}.
\]
This bijective correspondence allows to ``rename'' frames with transformations. Using this renaming, the detector $\psi$ can be rewritten as a function $\phi$ that outputs directly a transformation $\psi(\bx) = \phi(\bx)\bff_0$ instead of a frame.

With this substitution, the covariance constraint~\eqref{e:covariance} becomes
\begin{equation}\label{e:covariance1}
 \boxed{\phi(g\bx) \circ \phi(\bx)^{-1} \circ g^{-1} = 1}.	
\end{equation}
Note that, for the group of translations $G = T(2)$, this constraint corresponds directly to the objective function~\eqref{e:simple-loss}. Fig.~\ref{f:detector} provides two intuitive visualizations of this constraint. 

It is also useful to extend the learning objective~\eqref{e:simple-loss} as follows. As training data, we consider $n$ triplets $(\bx_i,\hat \bx_i, g_i), i=1,\dots,n$ comprising an image (patch) $\bx_i$, a transformation $g_i$, and the transformed and distorted image $\hat\bx_i = g \bx_i + \eta$. Here $\eta$ represents additive noise or some other useful distortion such as a random rescaling of the intensity which allows to train a more robust detector. The learning problem is then given by:
\begin{equation}\label{e:objective2}
    \min_\phi \frac{1}{n} \sum_{i=1}^n d(r_i,1)^2,
    \qquad
	r_i =  \phi(\hat\bx_i) \circ \phi(\bx_i)^{-1} \circ g_i^{-1}
\end{equation}
where $d(r_i,1)^2$ is the ``distance'' of the residual transformation $r_i$ from the identity.

\subsection{General covariant feature extraction}\label{s:partial}

\begin{figure}[t]
\begin{center}
	\includegraphics[height=0.13\textheight]{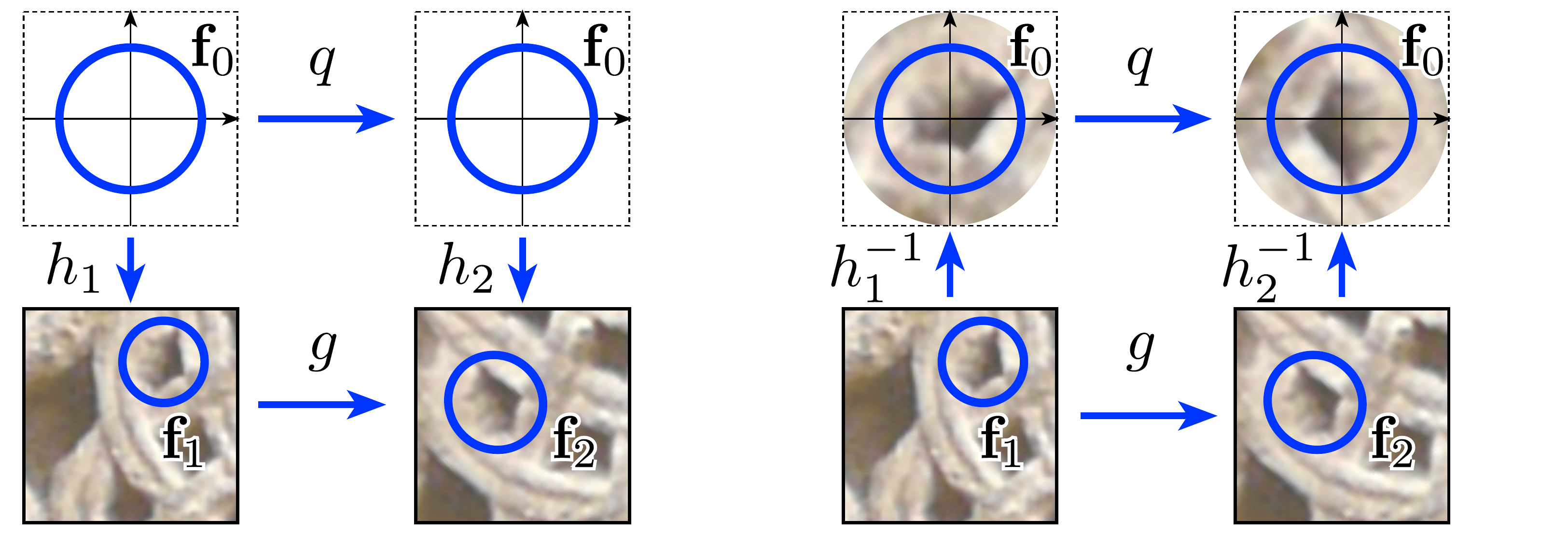}
\end{center}
\caption{\textbf{Left}: a (unoriented) circle identifies the translation and scale component of a similarity transformation $g\in G$, but leaves a residual rotation $q\in Q$ undetermined. Concretely, this could be the output of the SIFT detector prior orientation assignment. \textbf{Right:} normalization is achieved up to the residual transformation $q$.}\label{f:detector-quotient}
\end{figure}

The theory presented so far is insufficient to fully account for the properties of many common detectors. For this, we need to remove the assumptions that feature frames resolve (i.e. fix) completely the group of transformations $G$. Most detectors are in fact \emph{covariant with transformation groups larger than the ones that they can resolve}. For example, the Harris's detector is covariant with rotation and translation (in the sense that the same corners are extracted after the image is roto-translated), but, by detecting 2D points, it only resolves translations. Likewise, SIFT without orientation assignment is covariant to full similarity transformations but, by detecting circles, only resolves dilations (i.e.\ rotations remains undetermined; Fig.~\ref{f:detector-quotient}).

Next, we explain how eq.~\eqref{e:covariance1} must be modified to deal with detectors that (i) are covariant with a transformation group $G$ but (ii) resolve only a subgroup $H \subset G$. In this case, the detector function $\phi(\bx)\in H$ returns a transformation in the smaller group $H$, and the covariance constraint~\eqref{e:covariance1} is satisfied up to a complementary transformation $q \in Q$ that makes up for the part not resolved by the detector:
\begin{equation}\label{e:covariance-relaxed}
    \exists q \in Q : \quad \phi(g\bx) \circ q \circ \phi(\bx)^{-1} \circ g^{-1} = 1.
\end{equation}
This situation is illustrated graphically in Fig.~\ref{f:detector-quotient}.

For this construction to work, given $H \subset G$, the group $Q \subset G$ must be chosen appropriately. In eq.~\eqref{e:covariance-relaxed}, and following Fig.~\ref{f:detector-quotient}, call $h_1 =\phi(\bx)$ and $h_2 = \phi(g\bx)$. Rearranging the terms, we get that $h_2q = h_1g$, where $h_2\in H, q \in Q$ and $h_1g\in G$. This means that any element in $G$ must be expressible as a composition $hq$, i.e.\ $G = HQ= \{hq : h\in H,q\in Q\}$. Formally (proofs in appendix):

\begin{proposition}\label{p:one}
If the group $G = HQ$ is the product of the subgroups $H$ and $Q$, then, for any choice of $g\in G$ and $h_1\in H$, there is always a decomposition
\begin{equation}\label{e:dec}
		h_2 q h_1^{-1} g^{-1} = 1, \quad \text{such that}\quad h_2 \in H,\ q\in Q.
\end{equation}
\end{proposition}

In practice, given $G$ and $H$, $Q$ is usually easily found as the ``missing transformation''; however, compared to~\eqref{e:covariance}, the transformation $q$ in constraint~\eqref{e:covariance-relaxed} is an extra degree of freedom that complicates optimization. Fortunately, in many cases the following proposition shows that there is only one possible $q$:

\begin{proposition}\label{p:two}
If $H \triangleleft G$ is \emph{normal} in $G$ (i.e. $\forall g\in G, h \in H: g^{-1} h g \in H$) and $H \cap Q =\{1\}$, then, given $g \in G$, the choice of $q$ in the decomposition~\eqref{e:covariance-relaxed} is unique.
\end{proposition}

The next section works through several concrete examples to illustrate these concepts.

\subsection{A taxonomy of detectors}\label{s:taxonomy}

This section applies the theory developed above to standard detectors. Concretely, we limit ourselves to transformations up to affine, and write:
\[
\quad
h_i = \begin{bmatrix}
M_i & P_i \\
0 & 1
\end{bmatrix},\ %
q = \begin{bmatrix}
L & 0 \\
0 & 1
\end{bmatrix},\ %
g = \begin{bmatrix}
A & T \\
0 & 1
\end{bmatrix}.
\]
Here $P_i$ can be interpreted as the centre of the feature in image $\bx_i$ and $M_i$ as its affine shape, $(A,T)$ as the parameters of the image transformation, and $L$ as the parameter of the complementary transformation not fixed by the detector. The covariance constraint~\eqref{e:covariance-relaxed}  can be written, after a short calculation, as
\begin{equation}\label{e:explicit1}
M_2 L M_1^{-1} = A, \qquad P_2 - A P_1 = T.
\end{equation}
As a first example, consider a basic corner detector that resolves translations $H = G = T(2)$ with no (non-trivial) complementary transformation $Q=\{1\}$. Hence $M_1 = M_2 = L = A = I$ and \eqref{e:covariance-relaxed} becomes:
\begin{equation}\label{e:preharris}
    P_2 - P_1 = T.
\end{equation}
This is the same expression found in the simple example of Sect.~\ref{s:covconst} and requires the detected features to have the correct relative shift $T$.

The Harris corner detector is similar, but is covariant with rotations too. Formally, $H=T(2)\subset G=S\!E(2)$ (Euclidean transforms) and $Q=SO(2)$ (rotations). Since $T(2) \triangleleft S\!E(2)$ is a normal subgroup, we expect to find a unique choice for $q$. In fact, it must be $M_i = I$, $A = L = R$, and the constraint reduces to:
\begin{equation}\label{e:harris}
P_2 - R P_1 = T.
\end{equation}

In SIFT, $G=S(2)$ is the group of similarities, so that $A = sR$ is the composition of a rotation $R\in SO(2)$ and an isotropic scaling $s\in\real_+$. SIFT prior to orientation assignment resolves the subgroup $H$ of dilations (scaling and translation), so that $M_i = \sigma_i I$ (scaling) and the complement is a rotation $L \in SO(2)$. Once again $H \triangleleft G$, so the choice of $q$ is unique, and in particular $L = R$. The constraint reduces to:
\begin{equation}\label{e:sift}
P_2 - sR P_1 = T, \qquad \sigma_2/\sigma_1 = s.
\end{equation}
When orientation assignment is added to SIFT, the similarities are completely resolved $H=G=S(2)$, $M_i = \sigma_i R_i$ is a rotation and scaling, and the constraint becomes:
\begin{equation}\label{e:orsift}
P_2 - sR P_1 = T, \qquad \sigma_2/\sigma_1 = s, \qquad R_2R_1^\top = R.
\end{equation}

Affine detectors such as Harris-Affine (without orientation assignment) are more complex. In this case $G = A(2)$ are affinities and $H = U\!A(2)$ are \emph{upright affinities}, i.e. affinities where the linear map $M_i \in LT_+(2)$ is a lower-triangular matrix with positive diagonal (these affinities, which still form a group, leave the ``up'' direction unchanged). The residual $Q= SO(2)$ are rotations and $HQ = G$ is still satisfied. However, $U\!A(2)$ is \emph{not} normal in $A(2)$, Prop.~\ref{p:two} does not apply, and the choice of $Q$ is \emph{not} unique.\footnote{Concretely, from $M_2L = AM_1$ the complement matrix $L$ is given by the QR decomposition of the r.h.s.\ which is a function of $M_1$, i.e.\ not unique.} The constraint has the form:
\begin{equation}\label{e:harrisaff}
P_2 - A P_1 = T, \qquad M_2^{-1} A M_1 \in SO(2).
\end{equation}
For affine detectors with orientation assignment, $H = G = A(2)$ and the constraint is:
\begin{equation}\label{e:orharrisaff}
P_2 - A P_1 = T,\qquad M_2M_1^{-1} = A.
\end{equation}

The generality of our formulation allows learning many new types of detectors. For example, by setting $H=T(2)$ and $G=A(2)$ it is possible to train a corner detector such as Harris which is covariant to full affine transformations. Furthermore, a benefit of working with transformations instead of feature frames is that we can train detectors that would be difficult to express in terms of geometric primitives. For instance, by setting $H=SO(2)$ and $G=SE(2)$, we can train a \emph{orientation detector} which is covariant with \emph{rotation and translation}. As for affine upright features, in this case $H$ is not normal in $G$ so the complementary translation $q=(I,T')\in Q$ is not uniquely fixed by $g=(R,T)\in G$; nevertheless, a short calculation shows that the only part of~\eqref{e:covariance-relaxed} that matters in this case is
\begin{equation}\label{e:rotdet}
    R_2^\top R_1 = R
\end{equation}
where $h_i =(R_i,0)$ are the rotations estimated by the regressor.

\section{Implementation}\label{s:implementation}

This section discusses several implementation details of our method: the parametrization of transformations, example CNN architectures, multiple features detection, efficient dense detection, and preparing the training data.

\paragraph{Transformations: parametrization and loss.}\label{s:parametrization}

Implementing~\eqref{e:objective2} requires parametrizing the transformation $\phi(\bx) \in H$ predicted by the regressor. In the most general case of interest here, $H=A(2)$ are affine transformations and the simplest approach is to output the corresponding matrix of coefficients:
\[
	\phi(\bx)=
	\begin{bmatrix}
	\ba & \bb & \bp \\
	0 & 0 & 1 	
	\end{bmatrix}
	=
	\begin{bmatrix}
	a_u & b_u & p_u\\
	a_v & b_v & p_v\\
	0 & 0 & 1 	
	\end{bmatrix}.
\]
Here $\bp$ can be interpreted as the feature center and $\ba$ and $\bb$ as the feature affine shape. By rearranging the terms in~\eqref{e:covariance}, the loss function in~\eqref{e:objective2} takes the form
\begin{equation}\label{e:frob}
	d^2(r,1) = \min_{q\in Q} \| g \phi(\bx) - \phi(g \bx) q\|_F^2,
\end{equation}
where $\|\cdot\|_F$ is the Frobenius norm. As seen before, the complementary transformation $q$ is often uniquely determined given $g$ and the minimization can be removed by substituting this fixed value for $q$. In practice, $g$ and $q$ are also represented by matrices, as described in Sect.~\ref{s:taxonomy}.

When the resolved transformations $H$ are less general than affinities, the parametrization can be adjusted accordingly. For instance, for the basic detector of Sect.~\ref{s:covconst}, where $H=T(2)$, on can fix $\ba=(1,0)$, $\bb=(0,1)$, $q = I$ and $g = (I,T)$, which reduces to eq.~\eqref{e:simple-loss}. If, on the other hand, $H=SO(2)$ are rotation matrices as for the orientation detector~\eqref{e:rotdet},
\begin{equation}\label{e:parrot}
\phi(\bx) = \frac{1}{\sqrt{a_u^2+a_v^2}}
\begin{bmatrix}
	a_u & -a_v & 0\\
	a_v & a_u & 0 \\
	0 & 0 & 1
\end{bmatrix}.
\end{equation}

\paragraph{Network architectures.}\label{s:arch}

\begin{table}[t]
\centering
\caption{\emph{Network architectures.} The \archsmall and \archbig CNN architectures used which consist of a small number of convolutional layers applied densely and with no padding. The filter sizes and number is specified in the top part of each cell. Filters are followed by ReLU layers and, where indicated, by $2 \times 2$ max pooling and/or LRN.}\label{t:arch}
\vspace{0.5em}
\scriptsize
\begin{tabular}{|c||c|c|c|c|c|c|c| }
	\hline Model & Conv1 & Conv2 & Conv3 & Conv4 & Conv5 & Conv6 & Conv7  \\ \hline
	\multirow{2}{*}{\archsmall} & $5 \times 5 \times 40$ & $5 \times 5 \times 100$ & $4 \times 4 \times 300$  & $1 \times 1 \times 500$ & $1 \times 1 \times 500$  &  $1 \times 1 \times 2$ & \\ 
	& Pool $\downarrow 2$ & Pool $\downarrow 2$ & & & & & \\
	\hline
	\multirow{2}{*}{\archbig} & $5 \times 5 \times 60$ & $5 \times 5 \times 150$ & $4 \times 4 \times 450$  & $1 \times 1 \times 600$ & $1 \times 1 \times 600$  &  $1 \times 1 \times 600$ & $1 \times 1 \times 2$ \\ 
	& Pool $\downarrow 2$ & Pool $\downarrow 2$ + LRN & & & & & \\
	\hline 
\end{tabular} 
\end{table}

One of the benefits of our approach is that it allows to  use deep neural networks in order to implement the feature regressor $\phi(\bx)$. Here we experiment with two such architectures, \archsmall and \archbig, summarized in Tab.~\ref{t:arch}. For fast detection, these resemble the compact LeNet model of~\cite{lecun98gradient}. The main difference between the two is the number of layers and filters. The loss \eqref{e:frob} is differentiable and easily implemented in a network loss layer. Note that the loss requires evaluating the network $\phi$ twice, once applied to image $\bx$ and once to image $g \bx$. Like in siamese architectures, these can be thought of as two networks with shared weights.

When implemented in a standard CNN toolbox (in our case in MatConvNet \cite{vedaldi15matconvnet}), multiple patch pairs are processed in parallel by a single CNN execution in what is known as a minibatch. In practice, the operations in \eqref{e:frob} can be implemented using off-the-shelf CNN components. For example, the multiplication by the affine transformation $g$ in~\eqref{e:frob}, which depends on which pair of images in the batch is considered, can be implemented by using convolution routines, $1 \times 1$ filters, and so called ``filter groups''.

\begin{figure}[t]
	\centering
	\begin{tabular}{R{0.1\textwidth}  m{0.9\textwidth}}
		& {\bf \detnet} \\
		\rule{0pt}{4ex}  
		\rotatebox{90}{Train} & \includegraphics[width=0.9\linewidth]{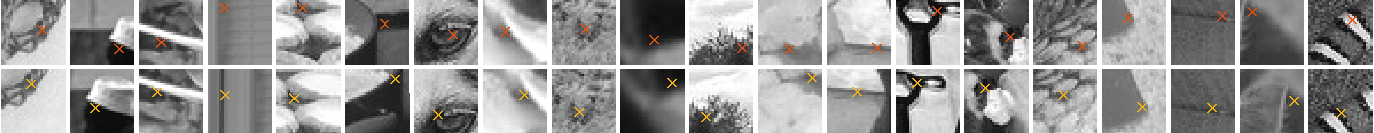} \\
		\rotatebox{90}{Easy} & \includegraphics[width=0.9\linewidth]{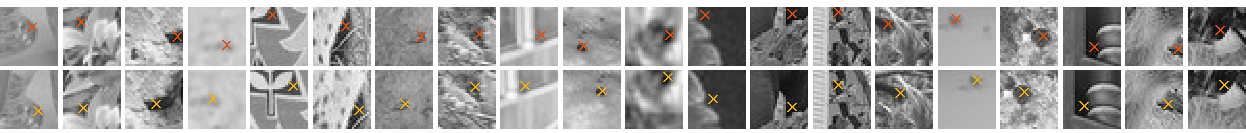} \\
		\rotatebox{90}{Hard} & \includegraphics[width=0.9\linewidth]{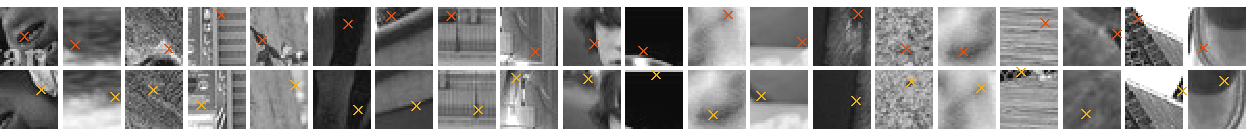} \\
		& {\bf \detnet} \\
		\rule{0pt}{4ex}  
		\rotatebox{90}{\centering Train} & \includegraphics[width=0.9\linewidth]{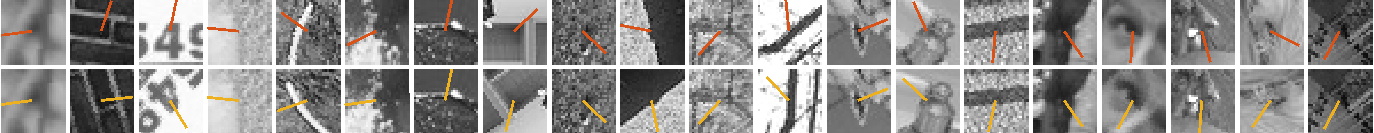} \\
		\rotatebox{90}{\centering Easy} & \includegraphics[width=0.9\linewidth]{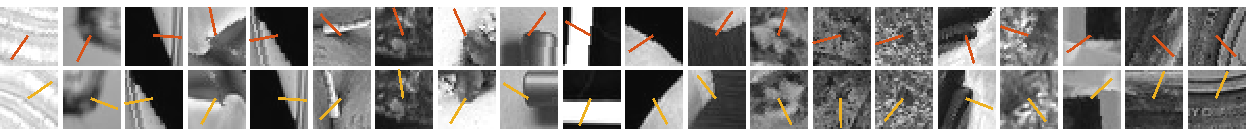} \\
		\rotatebox{90}{\centering Hard} & \includegraphics[width=0.9\linewidth]{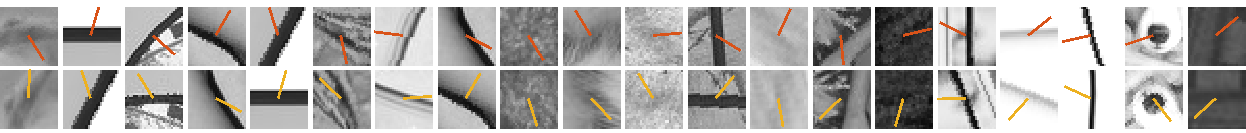} 
	\end{tabular}
\caption{\emph{Training and validation patches.} Example of training triplets $(\bx_1,\bx_2,g)$ ($\bx_1$ above and $\bx_2 = g\bx_1$ below) for different detectors. The figure also shows ``easy'' and ``hard'' patch pairs, extracted from the validation set based on the value of the loss~\eqref{e:parrot}. The crosses and bars represent respectively the detected translation and orientation, as learned by \detnet-L and \rotnet-L.}\label{f:training}
\end{figure}

\paragraph{From local regression to global detection.}\label{s:voting}

The formulation~\eqref{e:objective2} learns a function $\psi$ that maps an image patch $\bx$ to a single detected feature $\bff =\psi(\bx)$. In order to detect multiple features in a larger image, the function $\psi$ is simply applied convolutionally at all image locations (Fig.~\ref{f:spalsh}). Then, due to covariance, partially overlapping patches $\bx$ that contain the same feature are mapped by $\psi$ to the same detection $\bff$. Such duplicate detections are collapsed and their number, which reflects the stability of the feature, is used as detection confidence. 

For point features ($G=T(2)$), this voting process is implemented efficiently by accumulating votes in a map containing one bin for each pixel in the input image. Votes are accumulated using bilinear interpolation, after which non-maxima suppression is applied with a radius of two pixels. This scheme can be easily extended to more complex features, particularly under the reasonable assumption that only one feature is detected at each image location.

Note that some patches may in practice contain two or more clearly visible feature anchors. The detector $\psi$ must then decide which one to select. This is not a significant limitation at test time (as the missed anchors would likely be selected by a translated application of $\psi$). Its effect at training time is discussed later.

\paragraph{Efficient dense evaluation.}\label{s:efficient}

As most CNNs, architectures \archsmall and \archbig rapidly downsample their input for efficiency. In order to perform dense feature detection, the easiest approach is to reapply the CNNs to slightly shifted versions of the image, filling the ``holes'' left in the downsampled output. An equivalent but much more efficient method, which reuses significant computations in the denser early layers of the network, is the {\em \`a trous} algorithm~\cite{mallat08a-wavelet,papandreou15modeling}.

We propose here an algorithm equivalent to \emph{\`a trous} which is just as efficient and more easily implemented. Given a CNN layer $\bx_l = \phi_l(\bx_{l-1})$ that downsamples the input tensor $\bx_{l-1}$ by a factor of, say, two, the downsampling factor is changed to one, and the now larger output $\bx_l$ is split into four parts $\bx_l^{(k)},k=1,\dots,4$. Each part is obtained by downsampling $\bx_l$ after shifting it by zero or one pixels in the horizontal and vertical directions (for a total of four combinations). Then the deeper layers of the networks are computed as usual on the four parts independently. The construction is repeated whenever downsampling needs to be performed.

Detection speed can be improved with evaluating the regressor with stride 2 (at every second pixel). We refer to these detector as \emph{\detnet S2}. Source code and the \detnet models are freely available\footnote{\url{https://github.com/lenck/ddet}}.

\paragraph{Training data.}\label{s:training}

Training images are obtained from the ImageNet ILSVRC 2012 training data~\cite{russakovsky14imagenet}, extracting twenty random $57 \times 57$ crops per image, for up to $6M$ crops. Uniform crops are discarded since they clearly cannot contain any useful anchor. To do so, the absolute response of a LoG filter of variance $\sigma = 2.5$ is averaged and the crop is retained if the response is greater than 1.5 (image intensities are in the range $[0,255]$). Note that, combined with random selection, this operation \emph{does not} center crops on blobs or any other pre-defined anchors, but simply discards uniform or very low contrast crops.

Recall that the formulation Sect.~\ref{s:beyond} requires triplets $(\bx_1,\bx_2,g)$. A triplet is generated by randomly picking a crop and then by extracting $28 \times 28$ patches $\bx_1$ and $\bx_2$ within 20 pixels of the crop center (Fig.~\ref{f:training}). This samples two patches related by translation, corresponding to the translation sampled in $g$, while guaranteeing that patches overlap by least $27\%$. Then the linear part of $g$ is sampled at random and used to warp $\bx_2$ around its center. In order too achieve better robustness to photometric transformations, additive ($\pm 8\%$ of the intensity range) and multiplicative ($\pm 40\%$ of a pixel intensity) is added to the pixels .

Training uses batches of 64 patch pairs. An epoch contains $40 \cdot 10^3$ pairs, and the data is resampled after each epoch completes. The learning rate is set to $\lambda = 0.01$ and decreased tenfold when the validation error stops decreasing. Usually,  training converges after 60 epochs, which, due to the small size of the network and input patches, takes no more than a couple of minutes on a GPU.

\section{Experiments}\label{s:experiments}

We apply our framework to learn two complementary types of detectors in order to illustrate the flexibility of the approach: a corner detector (Sect.~\ref{s:corner}) and an orientation detector (Sect.~\ref{s:orientation}).

\paragraph{Evaluation benchmark and metrics.} We compare the learned detectors to standard ones: FAST~\cite{rosten06machine,rosten2005fusing} (using OpenCV's implementation\footnote{\url{opencv.org}}), the Difference of Gaussian detector (DoG) or SIFT~\cite{lowe04distinctive}, the Harris corner point detector \cite{harris88combined} and Hessian point detector \cite{mikolajczyk02affine} (all using VLFeat's implementation\footnote{\url{www.vlfeat.org}}). All experiments are performed at a single scale, but all detectors can be applied to a scale space pyramid if needed.

For evaluation of the corner detector, we use the standard VGG-Affine benchmark dataset~\cite{mikolajczyk02affine}, using both the \emph{repeatability} and \emph{matching score} criteria. For matching score, SIFT descriptors are extracted from a fixed region of $41 \times 41$ pixels around each corner. A second limitation in the original protocol of~\cite{mikolajczyk02affine} is that repeatability can be made arbitrarily large simply by detecting enough features. Thus, in order to control for the number of features detected, we compute repeatability and matching score as the feature detection threshold is increased; we then plot the metrics as functions of the number of feature selected in the first image.

VGG-Affine contains scenes related by homography. We also consider the more recent DTU-Robots dataset~\cite{aanaes12interesting} that contains 3D objects under changing viewpoint. Matches in DTU dataset are estimated using the known 3D shape of the objects and position of the camera. The data is divided in three ``arcs'', corresponding to three swipes of the robotic camera at different distances from the scene (0.5, 0.65, and 0.8m respectively). Due to the large number of images in this dataset, only aggregated results for $n=600$ are reported.

\begin{figure}[t]
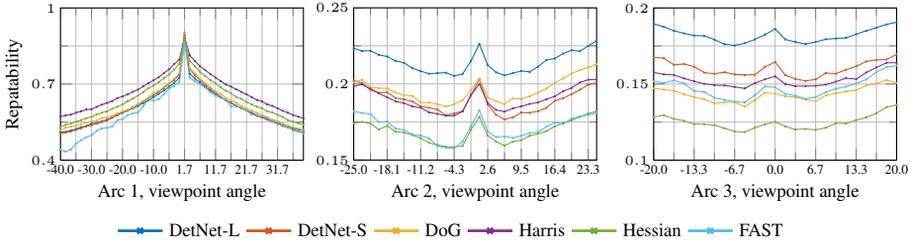

	\begin{center}
		\begin{tiny}
		\setlength{\figureheight}{2cm}
		\setlength{\figurewidth}{0.28\linewidth}
		\begin{tabular}{ c c c }
			\input{results/rep/dtu_arc1_rep.tkiz} &
			\input{results/rep/dtu_arc2_rep.tkiz} &
			\input{results/rep/dtu_arc3_rep.tkiz}
		\end{tabular}
	   	\begin{tikzpicture}
	   	\definecolor{mycolor1}{rgb}{0.00000,0.44700,0.74100}%
	   	\definecolor{mycolor2}{rgb}{0.85000,0.32500,0.09800}%
	   	\definecolor{mycolor3}{rgb}{0.92900,0.69400,0.12500}%
	   	\definecolor{mycolor4}{rgb}{0.49400,0.18400,0.55600}%
	   	\definecolor{mycolor5}{rgb}{0.46600,0.67400,0.18800}%
	   	\definecolor{mycolor6}{rgb}{0.30100,0.74500,0.93300}%
	   	\begin{customlegend}[legend columns=6,
	   	legend style={align=left,draw=none},
	   	legend cell align=left,
	   	legend entries={DetNet-L, DetNet-S, DoG, Harris, Hessian, FAST}]
	   	\addlegendimage{color=mycolor1,solid,line width=1.0pt,mark size=1.5pt,mark=x,mark options={solid}} 
	   	\addlegendimage{color=mycolor2,solid,line width=1.0pt,mark size=1.5pt,mark=x,mark options={solid}} 
	   	\addlegendimage{color=mycolor3,solid,line width=1.0pt,mark size=1.5pt,mark=x,mark options={solid}} 
	   	\addlegendimage{color=mycolor4,solid,line width=1.0pt,mark size=1.5pt,mark=x,mark options={solid}} 
	   	\addlegendimage{color=mycolor5,solid,line width=1.0pt,mark size=1.5pt,mark=x,mark options={solid}} 
	   	\addlegendimage{color=mycolor6,solid,line width=1.0pt,mark size=1.5pt,mark=x,mark options={solid}} 
	   	\end{customlegend}
	   	\end{tikzpicture}
	   	\vspace{-1em}
	   	\caption{\emph{Repeatability on the DTU Dataset} averaged over all 60 scenes, divided by arc. Repeatability is computed over the top 600 detections for each detector.}
	   	\label{fig:repmatch}
		\end{tiny}
	\end{center}
\end{figure}


\subsection{Corner or translation detector}\label{s:corner}

In this section we train a ``corner detector'' network \detnet. Using the formalism of Sect.~\ref{s:method}, this is a detector which is covariant with translations $G=T(2)$, corresponding to the covariance constraint of eq.~\eqref{e:simple-loss}. Fig.~\ref{f:training} provides a few examples of the patches used for training, as well as of the anchors discovered by learning.

Fig.~\ref{fig:repmatch} reports the performance of the two versions of \detnet, small and large, on the DTU data. As noted in~\cite{aanaes12interesting}, the Harris corner detector performs very well on the first arc; however, on the other two arcs \detnet-L clearly outperforms the other methods, whereas \detnet-S is on par or a little better than standard detectors.

\begin{figure}[t]
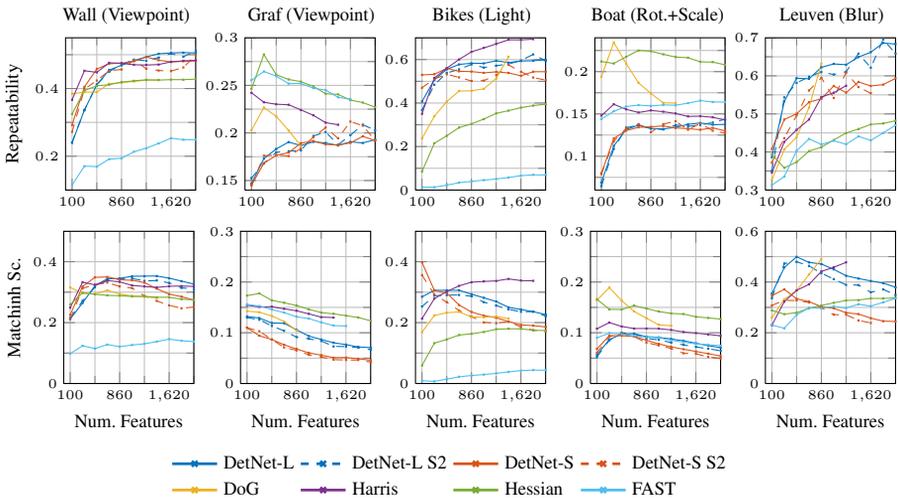

	\hspace{-2em}
	\begin{center}
		\begin{scriptsize}
			\setlength{\tabcolsep}{0.25pt}
			\setlength{\figureheight}{2cm}
			\setlength{\figurewidth}{0.15\linewidth}
			\pgfplotsset{
				tick label style={font=\tiny},
				label style={font=\scriptsize}}
			\newcommand{\RepRes}[2]{\input{results/rep/#1_#2.tkiz}}
			\begin{tabular}{ c c c c c }
				\hspace{3em}Wall (Viewpoint) & \hspace{2.5em}Graf (Viewpoint) &\hspace{2em}Bikes (Light) & \hspace{2.1em}Boat (Rot.+Scale) & \hspace{2em}Leuven (Blur) \\
				\input{results/rep/vgg_wall_rep.tkiz} & \input{results/rep/vgg_graf_rep.tkiz} & \input{results/rep/vgg_bikes_rep.tkiz} & \input{results/rep/vgg_boat_rep.tkiz} & \input{results/rep/vgg_leuven_rep.tkiz} \\
				\input{results/rep/vgg_wall_mscore.tkiz} & \input{results/rep/vgg_graf_mscore.tkiz} & \input{results/rep/vgg_bikes_mscore.tkiz} & \input{results/rep/vgg_boat_mscore.tkiz} & \input{results/rep/vgg_leuven_mscore.tkiz} \\
			\end{tabular}
			\vspace{-1em}
			\begin{center}
		   	\begin{tikzpicture}
			\definecolor{mycolor1}{rgb}{0.00000,0.44700,0.74100}%
			\definecolor{mycolor2}{rgb}{0.85000,0.32500,0.09800}%
			\definecolor{mycolor3}{rgb}{0.92900,0.69400,0.12500}%
			\definecolor{mycolor4}{rgb}{0.49400,0.18400,0.55600}%
			\definecolor{mycolor5}{rgb}{0.46600,0.67400,0.18800}%
			\definecolor{mycolor6}{rgb}{0.30100,0.74500,0.93300}%
		   	\begin{customlegend}[legend columns=4,
		   	legend style={align=left,draw=none},
		   	legend cell align=left,
		   	legend entries={DetNet-L, DetNet-L S2, DetNet-S, DetNet-S S2, DoG, Harris, Hessian, FAST}]
		   	\addlegendimage{color=mycolor1,solid,line width=1.0pt,mark size=1.5pt,mark=x,mark options={solid}} 
		    \addlegendimage{color=mycolor1,dashed,line width=1.0pt,mark size=1.5pt,mark=x,mark options={solid}} 
		    \addlegendimage{color=mycolor2,solid,line width=1.0pt,mark size=1.5pt,mark=x,mark options={solid}} 
   		    \addlegendimage{color=mycolor2,dashed,line width=1.0pt,mark size=1.5pt,mark=x,mark options={solid}} 
		   	\addlegendimage{color=mycolor3,solid,line width=1.0pt,mark size=1.5pt,mark=x,mark options={solid}} 
		   	\addlegendimage{color=mycolor4,solid,line width=1.0pt,mark size=1.5pt,mark=x,mark options={solid}} 
		   	\addlegendimage{color=mycolor5,solid,line width=1.0pt,mark size=1.5pt,mark=x,mark options={solid}} 
		   	\addlegendimage{color=mycolor6,solid,line width=1.0pt,mark size=1.5pt,mark=x,mark options={solid}} 
		   	\end{customlegend}
		   	\end{tikzpicture}
			\end{center}
		   	\end{scriptsize}
		   	\vspace{-1.5em}
			\caption{\emph{Repeatability and matching score on VGG dataset} comparing two versions of DetNet and standard detectors controlled for an increasing number of detected features. Dashed line values are for DetNet with stride 2. Scores are computed as an average over all 5 transformed images for each set (e.g.~``wall'').}
			\label{fig:repmatch2}
		\end{center}
\end{figure}

Fig.~\ref{fig:repmatch2} evaluates the method on the VGG-Affine dataset. Here the learned networks perform generally well too, outperforming existing detectors in some scenarios and performing less well on others. Note that our current implementation is the simplest possible and the very first of its kind; in the future more refined setups may improve the learned detectors across the board (Sect.~\ref{s:discussion}).

\begin{table}[t]
	\centering
	\caption{The detection speed (in FPS) for different image sizes of all tested detectors, computed as an average over 10 measurements. Please not that the \detnet detectors run on a GPU, other detectors run on a CPU.\label{tab:detspeed}}
	\vspace{0.5em}
	\begin{scriptsize}
		\begin{tabular}{| c |  r  r |  r  r |  r  r  r  r  |}\hline
			& \detnet-L& \detnet-L S2& \detnet-S& \detnet-S S2& Harris& DoG& Hessian& FAST \\ \hline 
			\textbf{$320 \times 240 $}  & 9.16  & 33.14  & 27.26  & 83.16  & 144.39  & 88.64  & 150.34  & 439.68  \\ 
			\textbf{$800 \times 600 $}  & 1.45  & 5.87  & 4.68  & 19.32  & 15.65  & 8.00  & 17.45  & 328.20  \\ 
			\textbf{$1024 \times 768 $}  & 0.39  & 1.56  & 2.78  & 11.68  & 12.21  & 6.05  & 11.17  & 206.96  \\ 
			\hline 
		\end{tabular}
	\end{scriptsize}

\end{table}

The speed of the tested detectors is shown in Table~\ref{tab:detspeed}. While our goal is not to obtain the fastest detector but rather to demonstrate the possibility of learning detectors from scratch, we note that even an unoptimised MATLAB implementation can achieve reasonable performance on a GPU, especially with stride 2 with a slightly decreased performance compared to the dense evaluation (see Figure~\ref{fig:repmatch2}).

\subsection{Orientation detector}\label{s:orientation}

\begin{figure}[t]
\begin{minipage}[c]{.6\textwidth}
	\begin{tiny}
		\pgfplotsset{legend style={font=\scriptsize}}
	\setlength{\figureheight}{1.8cm}
	\setlength{\figurewidth}{0.8\textwidth}
	\input{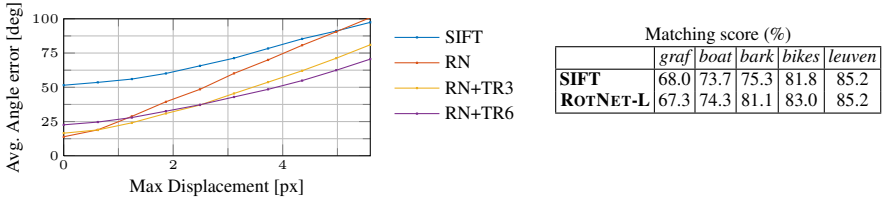}
	\end{tiny}
\end{minipage}%
\adjustbox{valign=b}{
\begin{minipage}[t]{.3\textwidth}
\centering
\scriptsize
\begin{tabular}{| l |  c | c |  c |  c |  c |}
\multicolumn{6}{c}{Matching score (\%)} \\ \hline
 & {\it graf}& {\it boat}& {\it bark}& {\it bikes}& {\it leuven} \\ \hline 
	\textbf{SIFT}      & 68.0 & 73.7 & 75.3 & 81.8 & 85.2 \\ 
	\textbf{\rotnet-L}  & 67.3 & 74.3 & 81.1 & 83.0 & 85.2 \\ 
	\hline 
\end{tabular}
\end{minipage}}
\caption{\emph{Orientation detector evaluation.} \textbf{Left}: versions of \rotnet (RN) and the SIFT orientation detector evaluated on recovering the relative rotation of random patch pairs. \textbf{Right}: matching score on the VGG-Affine benchmark when the native SIFT orientation estimation is replaced with \rotnet (percentage of correct matches using the DoG-Affine detector).}\label{fig:angleerr-sift}
\end{figure}

This section evaluates a network, \rotnet, trained for orientation detection. This detector resolves $H=SO(2)$ rotations and is covariant to Euclidean transformations $G = SE(2)$, which means that translations $Q=T(2)$ are nuisance factor that the detector should ignore. The corresponding form of the covariance constraint is given by eq.~\eqref{e:rotdet}. Training proceeds as above, using $28 \times 28$ pixels patches and, for $g$, random $2\pi$ rotations composed with a maximum nuisance translation of 0, 3, or 6 pixels, resulting in three different versions of the network (Fig.~\ref{f:training}).

The SIFT detector~\cite{lowe04distinctive} contains both a blob detector as well as an orientation detector, based on determining the dominant gradient orientation in the patch. Fig.~\ref{fig:angleerr-sift} compares the average angular registration error obtained by the SIFT orientation detector and different versions of \rotnet, measured from pairs of randomly-sampled image patches. We note that: 1) \rotnet is sensibly better than the SIFT orientation detector, with up to half the error rate, and that 2) while the error increases with the maximum nuisance translation between patches, networks that are trained to account for such translations are sensibly better than the ones that do not. Furthermore, when applied to the output of the SIFT blob detector, the improved orientation estimation results in an improved feature matching score, as measured on the VGG-Affine benchmark.

\section{Discussion}\label{s:discussion}

We have presented the first general machine learning formulation for covariant feature detectors. The latter is supported by a comprehensive theory of covariant detectors, and builds on the idea of casting detection as a regression problem. We have shown that this method can successfully learn corner and orientation detectors that outperform in several cases off-the-shelf detectors. The potential is significant; for example, the framework can be used to learn scale selection and affine adaptation CNNs. Furthermore, many significant improvements to our basic implementation are possible, including explicitly modelling detection strength/confidence, predicting multiple features in a patch, and jointly training detectors and descriptors. 


\paragraph{\textbf{Acknowledgements}}
We would like to thank ERC 677195-IDIU for supporting this research.

\appendix\section{Proofs}

\begin{proof}[of Proposition~\ref{p:one}] Due to group closure, $gh_1\in G$. Since $HQ=G$, then there must be $h_2\in H,q\in Q$ such that $h_2 q= g h_1$, and so $h_2 q h_1^{-1}g^{-1} = 1$.
\end{proof}

\begin{proof}[of Proposition~\ref{p:two}]
Let $h_2 q (h_1)^{-1} = h_2' q' (h_1')^{-1}$ be two such decompositions and multiply to the left by $(q)^{-1} (h_2')^{-1}$ and to the right by $h_1'$:
\[
	\underbrace{q^{-1}\ [(h_2')^{-1} h_2]\  q}_{\in H\text{\ (due to normality)}}
	\ %
	\underbrace{h_1^{-1} h_1'}_{\in H} = \underbrace{q^{-1} q'}_{\in Q}.
\]
Since this quantity is simultaneously in $H$ and in $Q$, it must be in the intersection $H \cap Q$, which by hypothesis contains only the identity. Hence $q^{-1}q' = 1$ and $q=q'$.
\end{proof}

{
\small
\bibliographystyle{ieee}
\bibliography{vedaldi-bibliography/bibliography,localbib}

\begin{thebibliography}{10}\itemsep=-1pt

\bibitem{aanaes12interesting}
H.~Aan{\ae}s, A.~Dahl, and K.~Steenstrup~Pedersen.
\newblock Interesting interest points.
\newblock {\em International Journal of Computer Vision}, pages 18--35, 2012.

\bibitem{baumberg00reliable}
A.~M. Baumberg.
\newblock Reliable feature matching across widely separated views.
\newblock In {\em Proc. {CVPR}}, pages 774--781, 2000.

\bibitem{beaudet78rotationally}
P.~R. Beaudet.
\newblock Rotationally invariant image operators.
\newblock In {\em International Joint Conference on Pattern Recognition},
  volume 579, page 583, 1978.

\bibitem{cordes11increasing}
K.~Cordes, B.~Rosenhahn, and J.~Ostermann.
\newblock Increasing the accuracy of feature evaluation benchmarks using
  differential evolution.
\newblock In {\em IEEE Symposium on Differential Evolution}, 2011.

\bibitem{dias95a-neural}
P.~Dias, A.~Kassim, and V.~Srinivasan.
\newblock A neural network based corner detection method.
\newblock In {\em {IEEE} Int. Conf. on Neural Networks}, 1995.

\bibitem{dufournaud99matching}
Y.~Dufournaud, C.~Schmid, and R.~Horaud.
\newblock Matching images with different resolutions.
\newblock In {\em Proc. {CVPR}}, 1999.

\bibitem{forstner86a-feature}
W.~F{\"o}rstner.
\newblock A feature based correspondence algorithm for image matching.
\newblock {\em International Archives of Photogrammetry and Remote Sensing},
  26(3):150--166, 1986.

\bibitem{freeman77a-corner-finding}
H.~Freeman and L.~S. Davis.
\newblock A corner-finding algorithm for chain-coded curves.
\newblock {\em IEEE Transactions on Computers}, (3):297--303, 1977.

\bibitem{guiducci88corner}
A.~Guiducci.
\newblock Corner characterization by differential geometry techniques.
\newblock {\em Pattern Recognition Letters}, 8(5):311--318, 1988.

\bibitem{han15matchnet:}
X.~Han, T.~Leung, Y.~Jia, R.~Sukthankar, and A.~C. Berg.
\newblock Matchnet: Unifying feature and metric learning for patch-based
  matching.
\newblock In {\em Proc. {CVPR}}, 2015.

\bibitem{harris88combined}
C.~Harris and M.~Stephens.
\newblock A combined corner and edge detector.
\newblock In {\em Proc. of The Fourth Alvey Vision Conference}, pages 147--151,
  1988.

\bibitem{holzer12learning}
S.~Holzer, J.~Shotton, and P.~Kohli.
\newblock Learning to efficiently detect repeatable interest points in depth
  data.
\newblock In {\em Proc. {ECCV}}, 2012.

\bibitem{kadir01saliency}
T.~Kadir and M.~Brady.
\newblock Saliency, scale and image description.
\newblock {\em Int. J. Computer Vision}, 45:83--105, 2001.

\bibitem{kienzle06learning}
W.~Kienzle, F.~A. Wichmann, B.~Sch{\"o}lkopf, and M.~O. Franz.
\newblock Learning an interest operator from human eye movements.
\newblock In {\em {CVPR} Workshop}, 2006.

\bibitem{lecun98gradient}
Y.~Lecun, L.~Bottou, Y.~Bengio, and P.~Haffner.
\newblock Gradient-based learning applied to document recognition.
\newblock {\em Proceedings of the IEEE}, Nov 1998.

\bibitem{lindeberg94scale-space}
T.~Lindeberg.
\newblock {\em Scale-Space Theory in Computer Vision}.
\newblock Springer, 1994.

\bibitem{lindeberg98feature}
T.~Lindeberg.
\newblock Feature detection with automatic scale selection.
\newblock {\em {IJCV}}, 30(2):77--116, 1998.

\bibitem{lindeberg94shape-adapted}
T.~Lindeberg and J.~Garding.
\newblock Shape-adapted smoothing in estimation of {3-D} depth cues from affine
  distortions of local {2-D} brightness structure.
\newblock In {\em Proc. {ECCV}}, 1994.

\bibitem{lowe99object}
D.~G. Lowe.
\newblock Object recognition from local scale-invariant features.
\newblock In {\em Proc. {ICCV}}, 1999.

\bibitem{lowe04distinctive}
D.~G. Lowe.
\newblock Distinctive image features from scale-invariant keypoints.
\newblock {\em {IJCV}}, 2(60):91--110, 2004.

\bibitem{mallat08a-wavelet}
S.~Mallat.
\newblock {\em A Wavelet Tour of Signal Processing}.
\newblock Academic Press, 2008.

\bibitem{matas02local}
J.~Matas, S.~Obdrz{\'a}lek, and O.~Chum.
\newblock Local affine frames for wide-baseline stereo.
\newblock In {\em Intl. Conference on Pattern Recognition}, 2002.

\bibitem{mikolajczyk01harris-laplace}
K.~Mikolajczyk and C.~Schmid.
\newblock Indexing based on scale invariant interest points.
\newblock In {\em Proc. {ICCV}}, 2001.

\bibitem{mikolajczyk02affine}
K.~Mikolajczyk and C.~Schmid.
\newblock An affine invariant interest point detector.
\newblock In {\em Proc. {ECCV}}, pages 128--142. Springer-Verlag, 2002.

\bibitem{olague11evolutionary-computer-assisted}
G.~Olague and L.~Trujillo.
\newblock Evolutionary-computer-assisted design of image operators that detect
  interest points using genetic programming.
\newblock {\em Image and Vision Computing}, 2011.

\bibitem{papandreou15modeling}
G.~Papandreou, I.~Kokkinos, and P.-A. Savalle.
\newblock Modeling local and global deformations in deep learning: Epitomic
  convolution, multiple instance learning, and sliding window detection.
\newblock {\em Proc. {CVPR}}, 2015.

\bibitem{paulin2015local}
M.~Paulin, M.~Douze, Z.~Harchaoui, J.~Mairal, F.~Perronin, and C.~Schmid.
\newblock Local convolutional features with unsupervised training for image
  retrieval.
\newblock In {\em {ICCV}}, 2015.

\bibitem{rohr92recognizing}
K.~Rohr.
\newblock Recognizing corners by fitting parametric models.
\newblock {\em {IJCV}}, 9(3), 1992.

\bibitem{rosenfeld73angle}
A.~Rosenfeld and E.~Johnston.
\newblock Angle detection on digital curves.
\newblock {\em Computers, IEEE Transactions on}, 100(9):875--878, 1973.

\bibitem{rosten2005fusing}
E.~Rosten and T.~Drummond.
\newblock Fusing points and lines for high performance tracking.
\newblock In {\em {ICCV}}, volume~2, 2005.

\bibitem{rosten06machine}
E.~Rosten and T.~Drummond.
\newblock Machine learning for high-speed corner detection.
\newblock In {\em Proc. {ECCV}}, 2006.

\bibitem{rosten10faster}
E.~Rosten, R.~Porter, and T.~Drummond.
\newblock Faster and better: a machine learning approach to corner detection.
\newblock In {\em {PAMI}}, volume~32, 2010.

\bibitem{russakovsky14imagenet}
O.~Russakovsky, J.~Deng, H.~Su, J.~Krause, S.~Satheesh, S.~Ma, Z.~Huang,
  A.~Karpathy, A.~Khosla, M.~Bernstein, A.~C. Berg, and L.~Fei-Fei.
\newblock Imagenet large scale visual recognition challenge, 2014.

\bibitem{sankar78a-parallel}
P.~Sankar and C.~Sharma.
\newblock A parallel procedure for the detection of dominant points on a
  digital curve.
\newblock {\em Computer Graphics and Image Processing}, 7(3):403--412, 1978.

\bibitem{schaffalitzky01viewpoint}
F.~Schaffalitzky and A.~Zisserman.
\newblock Viewpoint invariant texture matching and wide baseline stereo.
\newblock In {\em Proc. {ICCV}}, 2001.

\bibitem{schmid97local}
C.~Schmid and R.~Mohr.
\newblock Local greyvalue invariants for image retrieval.
\newblock {\em Pattern Analysis and Machine Intelligence, {IEEE} Transactions
  on}, 1997.

\bibitem{simo2015discriminative}
E.~Simo-Serra, E.~Trulls, L.~Ferraz, I.~Kokkinos, P.~Fua, and F.~Moreno-Noguer.
\newblock Discriminative learning of deep convolutional feature point
  descriptors.
\newblock In {\em {ICCV}}, 2015.

\bibitem{smith95SUSAN}
S.~M. Smith and J.~M. Brady.
\newblock Susan -- a new approach to low level image processing.
\newblock Technical report, Oxford University, 1995.

\bibitem{sochman09learning}
J.~Sochman and J.~Matas.
\newblock Learning fast emulators of binary decision processes.
\newblock {\em {IJCV}}, 2009.

\bibitem{triggs04detecting}
B.~Triggs.
\newblock Detecting keypoints with stable position, orientation, and scale
  under illumination changes.
\newblock In {\em Proc. {ECCV}}, 2004.

\bibitem{trujillo06synthesis}
L.~Trujillo and G.~Olague.
\newblock Synthesis of interest point detectors through genetic programming.
\newblock In {\em Proc. {GECCO}}, 2006.

\bibitem{tuytelaars00wide}
T.~Tuytelaars and L.~Van~Gool.
\newblock Wide baseline stereo matching based on local, affinely invariant
  regions.
\newblock In {\em Proc. {BMVC}}, pages 412--425, 2000.

\bibitem{vedaldi15matconvnet}
A.~Vedaldi and K.~Lenc.
\newblock Matconvnet -- convolutional neural networks for matlab.
\newblock In {\em Proc. {ACM} Int. Conf. on Multimedia}, 2015.

\bibitem{Yi16Learning}
K.~M. Yi, Y.~Verdie, P.~Fua, and V.~Lepetit.
\newblock {Learning to Assign Orientations to Feature Points}.
\newblock In {\em {CVPR}}, 2016.

\bibitem{Zagoruyko2015}
S.~Zagoruyko and N.~Komodakis.
\newblock Learning to compare image patches via convolutional neural networks.
\newblock In {\em {CVPR}}, 2015.

\bibitem{zbontar15computing}
J.~Zbontar and Y.~LeCun.
\newblock Computing the stereo matching cost with a convolutional neural
  network.
\newblock In {\em Proc. {CVPR}}, 2015.

\bibitem{zuliani05mathematical}
M.~Zuliani, C.~Kenney, and B.~S. Manjunath.
\newblock A mathematical comparison of point detectors.
\newblock In {\em Proc. {CVPR}}, 2005.

\end{thebibliography}
}
\end{document}